\documentclass[sigconf]{acmart}
\usepackage[tight,footnotesize]{subfigure}
\usepackage{float}
\usepackage{graphicx}
\usepackage{xcolor,colortbl}
\usepackage{tcolorbox}
\usepackage[nolist,nohyperlinks]{acronym}
\usepackage[english]{babel}
\usepackage[utf8]{inputenc}
\usepackage{bm}
\usepackage{lipsum}
\usepackage[nolist,nohyperlinks]{acronym}
\usepackage[inline]{enumitem}
\usepackage{url}

\graphicspath{ {./} }
\include{misc_macros}

 \renewcommand\footnotetextcopyrightpermission[1]{}

\newcommand{\ourmethod}{\textit{mmPhase}}

\begin{document}

\title{Poster: Dynamic Ego-Velocity estimation Using Moving mmWave Radar: A Phase-Based Approach}

\author{Argha Sen$^*$, Soham Chakraborty$^*$, Soham Tripathy$^*$, Sandip Chakraborty}
\affiliation{%
	\institution{Indian Institute of Technology Kharagpur, India}
	}
 \thanks{$^*$All the authors have contributed equally}
 
\email{arghasen10@gmail.com, sohamc1909@kgpian.iitkgp.ac.in, sohamtripathy2001@gmail.com, sandipc@cse.iitkgp.ac.in}


\begin{abstract}
    Precise ego-motion measurement is crucial for various applications, including robotics, augmented reality, and autonomous navigation. In this poster, we propose \ourmethod{}, an odometry framework based on single-chip millimetre-wave (mmWave) radar for robust ego-motion estimation in mobile platforms without requiring additional modalities like the visual, wheel, or inertial odometry. \ourmethod{} leverages a phase-based velocity estimation approach to overcome the limitations of conventional doppler resolution. For real-world evaluations of \ourmethod{} we have developed an ego-vehicle prototype. Compared to the state-of-the-art baselines, \ourmethod{} shows superior performance in ego-velocity estimation.
\end{abstract}
\maketitle

\section{Introduction}
Understanding the movement of mobile agents, needed in autonomous navigation or augmented reality settings, is crucial for perception and interaction. Ego-motion estimation, unlike map-based localization, doesn't rely on prior knowledge of the environment or infrastructure. Instead, it analyzes sensory data from the agent's movement to determine its position and orientation over time. Due to easier affordability and widespread availability, MEMS inertial sensors (IMUs) are commonly used for ego-motion estimation in various mobile platforms. However, their accuracy is limited by noise and bias, leading to significant drift in inertial odometry. To overcome these limitations, multi-modal odometry systems have been proposed, which combine inertial information with other modalities, such as visual or ranging information. However, the performance of Visual-Inertial Odometry (VIO) can degrade in challenging lighting conditions. Similar visibility issues affect LiDAR-Inertial Odometry (LIO), particularly when dealing with airborne obscurants like dust, fog, and smoke. LiDARs, while effective, are often bulky, heavy, and expensive compared to cameras, making them more common in high-end robotics rather than micro-robots or wearable devices.

This poster proposes ego-motion estimation using Commercial-Off-The-Shelf (COTS) mmWave radar to explore cost-effective alternatives to optical systems such as LiDAR. mmWave modality offers advantages over vision-based systems, particularly in robustness to environmental conditions such as scene illumination and airborne obscurants. Unlike LiDAR or mechanically scanning radar, it uses electronic beamforming, making it lightweight and suitable for micro-robots and mobile or wearable devices. Smartphones like the Google Pixel 4 and commercial drones already use mmWave radar for motion sensing and obstacle detection, making it a next generation pervasive sensing solution. 

\begin{figure}
    \centering
    \includegraphics[width=0.45\textwidth]{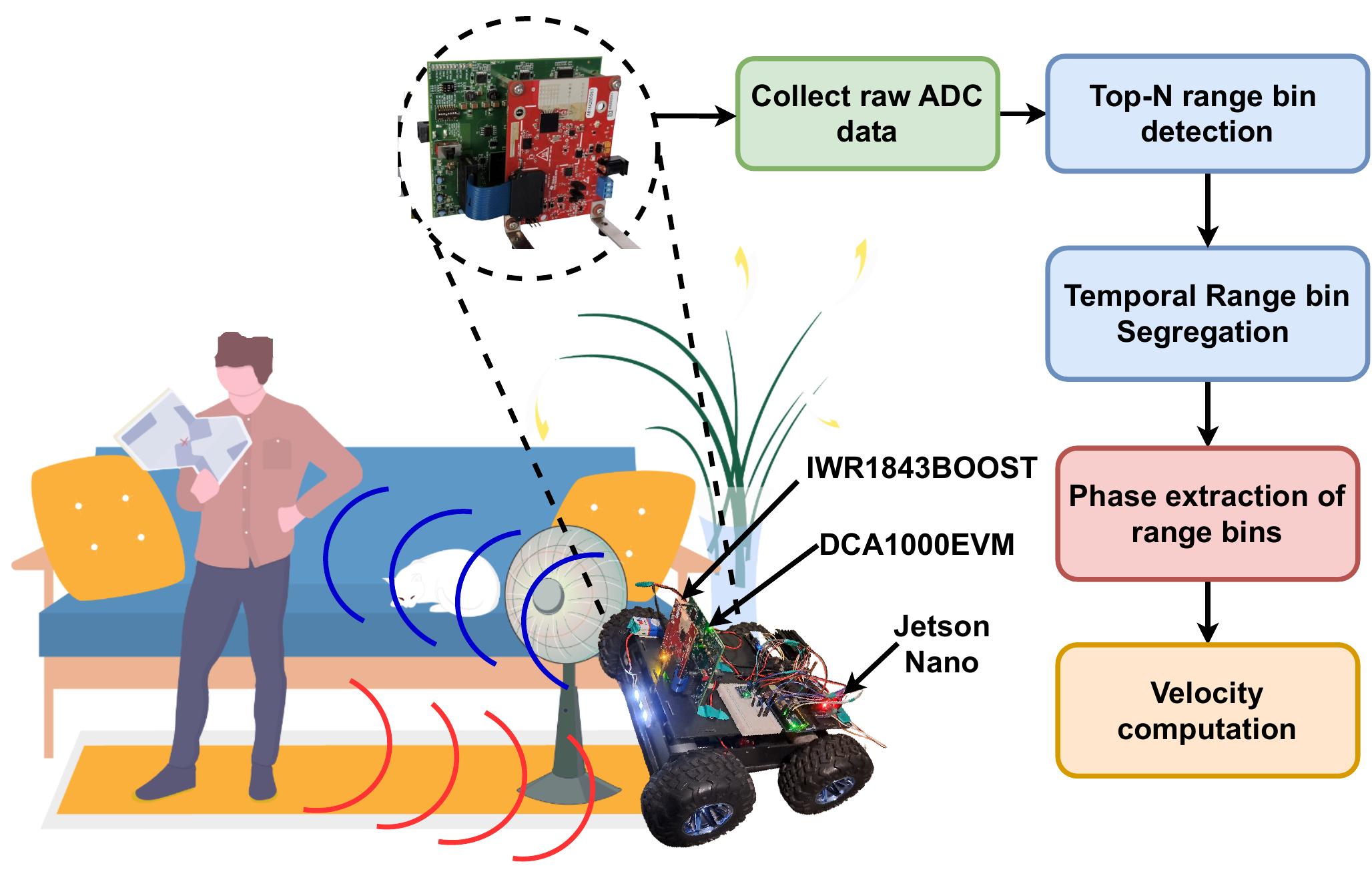}
    \caption{System overview of \ourmethod{}}
    \label{fig:sysOverview}
\end{figure}
The disadvantage of utilizing mmWave radar for indoor odometry lies in its inherent hardware limitations, resulting in sparse pointclouds with restricted angular resolution, susceptibility to noise from specular reflections, and significant multipath effects. This low-quality data makes conventional LiDAR-based methods, such as Iterative Closest Point (ICP), ineffective when applied directly to mmWave radar data. Also existing methods typically use range-doppler computation on the mmWave data, which has limitations in velocity resolution as the minimum doppler resolution of these radars is approximately $3.41$ cm/s~\cite{li2021signal}, restricting their ability to capture movements below this threshold. While some previous works~\cite{lu2020milliego,kramer2020radar,almalioglu2020milli} have employed multimodal approach by fusing mmWave radar data with other sensors, such as IMUs and RGB cameras, the potential of mmWave radar to complement these modalities remains uncertain. Additionally, incorporating recent advances in deep neural networks for visual or LiDAR odometry poses challenges due to heavy computational load, which may limit their use in mobile, wearable, and other resource-constrained devices. 

To address these challenges, we propose \ourmethod{}, that utilizes a phase-based velocity estimation method to overcome the inherent limitations of conventional doppler resolution. This approach enables to achieve low latency on embedded platforms compared to its DNN counterparts. We have developed a real-time prototype implementation of \ourmethod{} and conducted extensive real-world evaluations with several baselines. 

\section{Methodology}
\figurename~\ref{fig:sysOverview} summarizes the system overview of \ourmethod{}. On the collected raw ADC data from the mmWave radar, \ourmethod{} first applies a range-FFT and selects top $N$ peak values to isolate the range bins where potential reflectors are present. These $N$ range peaks can vary over the frames due to dynamic subject movement or due to multipath reflections. Over the frames, it checks if the reflector range bin is consistent (within $\pm 3$ range bins) to segregate the static reflectors from the noisy range peaks. Then, it collects the corresponding phase values from the selected range bins representing individual objects. Phase unwrapping is done on the collected phase values to make the phase values continuous. 

\begin{figure}
    \centering
    \subfigure[]{
    \includegraphics[width=0.20\textwidth]{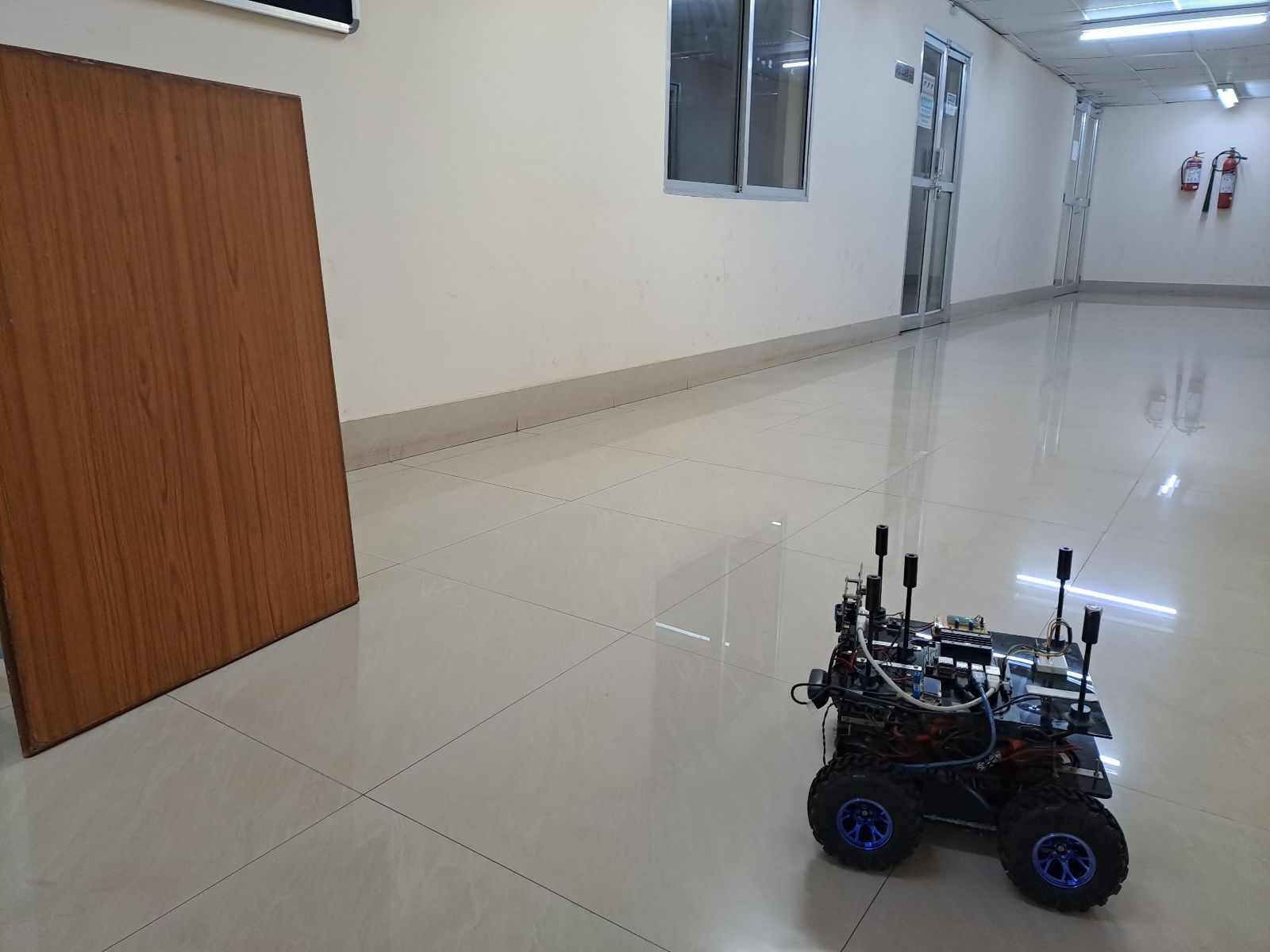}
    \label{fig:hardware}
    }\hfill
    \subfigure[]{
    \includegraphics[width=0.25\textwidth]{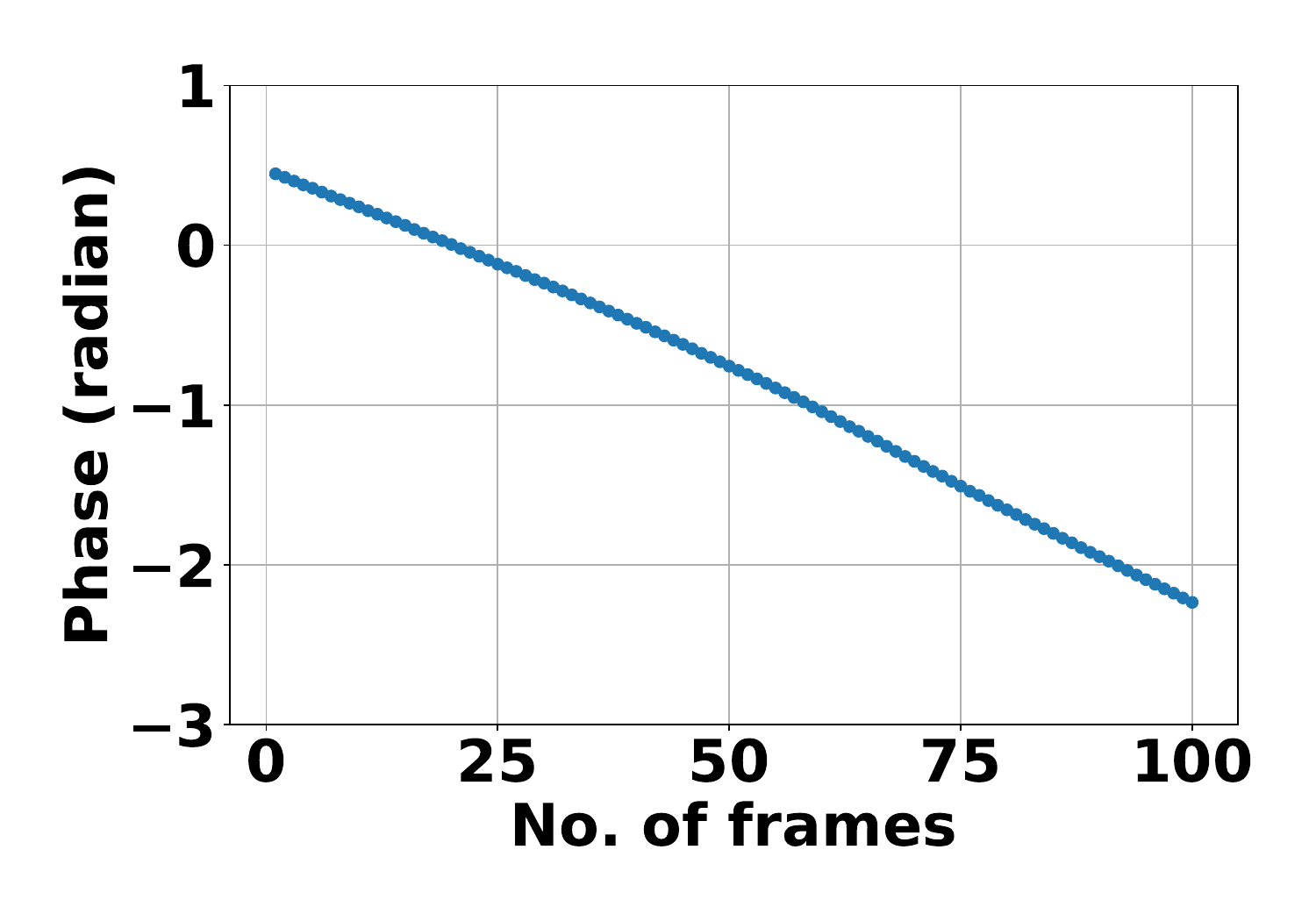}
    \label{fig:phase}
    }
    \caption{(a) \ourmethod{} hardware setup, (b) phase variation with the number of frames.}\label{fig:static_scene}
\end{figure}

The relation between the phase of the reflecting object and the distance at which the object is located can be given as $\phi = \frac{4\pi d}{\lambda}$. Thus, phase has a proportional relation to distance as can be observed from \figurename~\ref{fig:static_scene}, where we kept a single static object before the ego-vehicle and collected the raw mmWave data. As shown in \figurename~\ref{fig:phase}, the phase values decrease with time as the ego-vehicle moves towards the static object. The velocity of the ego-vehicle ($v_b$) for a static object can be represented as $\frac{d\phi}{dt} = \frac{4\pi v_b}{\lambda}$. From the phase values collected at a granularity ($dt$) of $T_c$ (chirp time) $+T_p$ (process time) $= 86 \mu$ secs, we can estimate the relative velocity of the ego-vehicle ($v_b$) from the above relation. As the $\lambda$ is in the millimetre range, a typical  $d \phi$ of around $0.057^o$ can capture the velocity at a granularity of $1.23$ cm/s~\cite{basak2022mmspy} compared to standard doppler based approach which can only work at a fixed doppler resolution (typically 3.41 cm/s). 

\section{Evaluation and Conclusion}
In \ourmethod{} setup, we use AWR1843BOOST EVM mmWave radar connected to a DCA1000EVM for fast data collection (5 FPS) and stored the data using a Jetson Nano mounted over the ego-vehicle. For software, we use Texas Instruments' mmWave Studio for radar configuration and PlatformIO Arduino IDE for uploading firmware to bot control. We collected the trajectory of our ego-vehicle using the Vicon Vero tracker (v1.3X). From the given trajectory, we computed the ground truth velocity of our ego-vehicle.  

\begin{figure}
    \centering
    \subfigure[]{
    \includegraphics[width=0.22\textwidth]{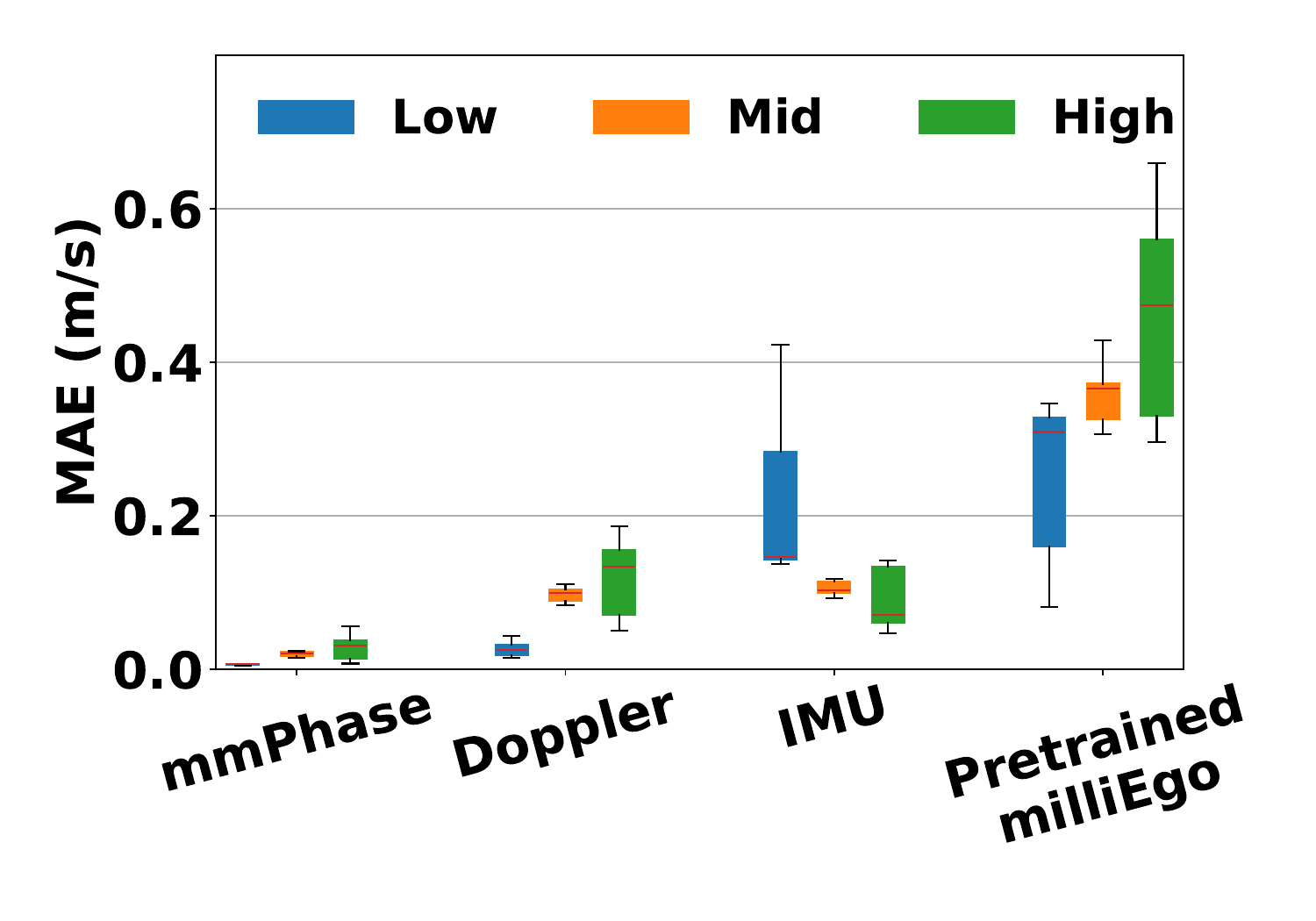}
    \label{fig:bp1}
    }\hfil
    \subfigure[]{
    \includegraphics[width=0.22\textwidth]{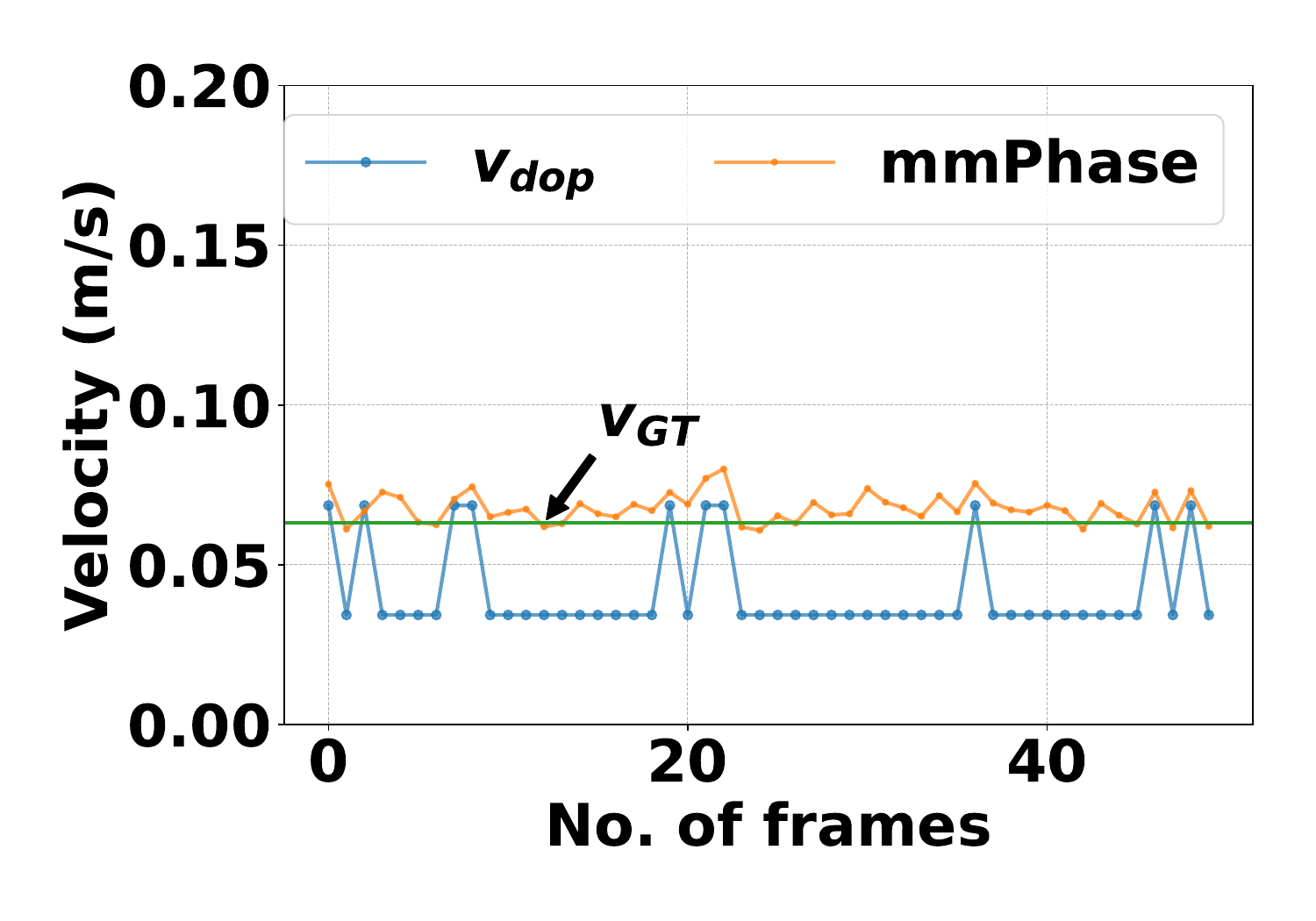}
    \label{fig:tracking}
    }
    \caption{(a) Mean Absolute Error in velocity estimation with respect to baselines, (b) Variation in the estimated velocity w.r.t. doppler-based approach at lower velocities.}
    \label{fig:box}
\end{figure}

We have compared our method with several baselines: (i) Doppler-based approach, which computes Doppler-FFT on the raw ADC data and estimates the velocity from the peak doppler bins, (ii) IMU-based odometry using a MPU6050, mounted over the ego-vehicle and (iii) Pre-trained milliEgo~\cite{lu2020milliego} model which fuses both mmWave range-angle heatmaps along with IMU to predict position and orientation. Our modified milliEgo returns the velocity directly instead of the positional coordinates. As shown in \figurename~\ref{fig:bp1}, the mean absolute error of our method is $4 \times$ lesser than its closest baseline, i.e., the doppler-based approach. As the velocity of the ego-vehicle increases, we have a higher error rate, primarily because at higher speeds, the phase component of the reflectors can get more noisy, and applying doppler-FFT can adversely affect the velocity estimation. Interestingly, IMU-based odometry exhibits superior performance at higher speeds than lower ones. During each data collection session covering the same distance, lower speeds necessitate more time, thereby leading to time drift issues within the IMU data.  The pre-trained milliEgo performs the worst as it is pre-trained in a different environment and suffers from both modalities' (IMU and mmWave) negative side. At velocities lower than the doppler resolution, the closest baseline, i.e., the doppler-based approach, suffers the most, as it's unable to capture sub-doppler movements accurately shown in \figurename~\ref{fig:tracking}. 

Unlike the existing methods, \ourmethod{} relies solely on mmWave raw phase data, overcoming not only the limitations such as sparse point clouds, and low-velocity resolutions but also offering low latency, making it suitable for resource-constrained mobile or wearable devices. In future, we intend to validate our setup further by incorporating multiple static and dynamic objects, different occlusions, and room settings. Also, we are considering a novel approach to ego-velocity estimation by employing a Physics-Informed Neural Network~\cite{raissi2019physics} to estimate the velocity. The knowledge of general physical laws between phase and velocity can help train neural networks (NNs) as regularization agents.

\bibliographystyle{plain}
\bibliography{main.bib}

\end{document}